\documentclass{article}

\usepackage[nonatbib, preprint]{neurips_2025}

\usepackage[utf8]{inputenc} 
\usepackage[T1]{fontenc}    
\usepackage{hyperref}       
\usepackage{url}            
\usepackage{booktabs}       
\usepackage{amsfonts}       
\usepackage{nicefrac}       
\usepackage{microtype}      
\usepackage{xcolor}         

\usepackage{macros}
\usepackage[square,numbers]{natbib}
\usepackage{float}
\usepackage{multicol}

\graphicspath{ {./img/} }
\setcitestyle{numbers}
\title{Basis Transformers for Multi-Task Tabular Regression}

\author{%
  Wei Min Loh$^{1,2}$ \quad Jiaqi Shang$^1$ \quad Pascal Poupart$^{1,2}$\\
  $^1$ University of Waterloo, $^2$ Vector Institute\\
  \texttt{\{wmloh,jiaqi.shang,ppoupart\}@uwaterloo.ca}
}

\begin{document}

\maketitle

\begin{abstract}
  Dealing with tabular data is challenging due to partial information, noise, and heterogeneous structure. Existing techniques often struggle to simultaneously address key aspects of tabular data such as textual information, a variable number of columns, and unseen data without metadata besides column names. We propose a novel architecture, \textit{basis transformers}, specifically designed to tackle these challenges while respecting inherent invariances in tabular data, including hierarchical structure and the representation of numeric values. We evaluate our design on a multi-task tabular regression benchmark, achieving an improvement of 0.338 in the median $R^2$ score and the lowest standard deviation across 34 tasks from the OpenML-CTR23 benchmark. Furthermore, our model has five times fewer parameters than the best-performing baseline and surpasses pretrained large language model baselines -- even when initialized from randomized weights.
\end{abstract}

\section{Introduction}


Our paper presents work for tabular, multi-modal (numerical values, text, categories, missing values), and multi-task regression. In contemporary tasks (in healthcare \cite{borisov2022deep}, finance, robotics \cite{novak2015survey}, etc.), there is often a multitude of data sources associated with different modalities that need to be leveraged simultaneously to achieve the best predictions possible.  Hence, traditional regression techniques that assume only numerical and categorical values are inadequate.  Missing values are also ubiquitous in real-world tasks and often present challenges that are handled heuristically by preprocessing and imputation techniques \cite{ccetin2022comprehensive}. Instead, it is desirable to work with techniques that naturally handle missing data out of the box.

In many areas, multi-task capabilities are desirable because the core process generates tabular data of similar formats but will be used in different downstream tasks. This enables a model to leverage information across different channels, encouraging transfer learning. Periodic minor changes to the format, such as an additional column, would not render an existing model unusable. Furthermore, metadata such as column names provides useful information that can facilitate transfer learning across tasks with similar attributes \cite{van2024tabular}.  Traditional tabular regression techniques often ignore such information, fail to transfer knowledge across tasks, and require a lot more training data.

Traditionally, tabular regression techniques assume that numerical values are normalized in some standard range for stability and numerical reasons, including avoiding exploding and vanishing gradients. However, such normalization reduces transfer learning opportunities across tasks where values in different ranges should not necessarily become similar. Furthermore, in zero (and few) shot learning scenarios, data normalization is simply impossible without prior knowledge or observing sufficiently many values that reveal the range of an attribute.  Hence, there is a need for regression techniques that preserve the scale of numerical values while still ensuring stable and effective learning.

\subsection{Desiderata}\label{sec:principles}

Ideally, techniques for tabular multi-modal multi-task regression should satisfy several desiderata in order to work with varying sources of data and handle the challenges described above.  We build upon the four desiderata by \citet{van2024tabular} which stipulate that tabular regression techniques should: (\setword{D1}{d1}) \textit{Handle entry values that can be numeric, categorical, textual or missing values}, (\setword{D2}{d2}) \textit{Handle variable number of columns}, (\setword{D3}{d3}) \textit{Leverage column names or other metadata}, (\setword{D4}{d4}) \textit{Be invariant to column order}. We propose two additional desiderata, whereby techniques should:

(\setword{D5}{d5}) \textit{Preserve and exploit tabular structure.}

Each table contains many rows; each row contains several entries (columns); each entry could contain a sequence (e.g. text). This separation should remain in the modelling process so that the inductive bias of the algorithm matches the nature of the data. Otherwise, flattening each row into an unstructured sequence of tokens ignores the explicit attribute-value pair structure of a table.

(\setword{D6}{d6}) \textit{Preserve the scale and precision of numeric values.}

As explained above, data normalization is not possible in zero-shot learning and mitigates transfer learning across multiple tasks where values in different ranges should not be made similar.

\subsection{Contributions}

We present \textit{basis transformers}, crafted to respect the invariances of structured modality. They are the product of fusing key ideas on: a good numeric representation compatible with text, and precise mixtures of information across tabular structures. This design fulfills all six desiderata. In this paper, our contributions include:

\vspace{-5pt}

\begin{enumerate}[leftmargin=0.5cm,label=(\arabic*)]
	\item Integrating sign-magnitude numeric representation into transformer-based architectures
	\item Transforming a regression problem into a multi-label classification problem, benefitting from scale-invariance training
	\item Designing a scalable architecture for tabular modality that abides by important tabular properties
	\item Introducing a loss reweighing scheme for regression models to focus on hard examples
	\item Demonstrating that the basis transformer has the best performance in multi-task tabular regression on the OpenML-CTR23 benchmark consisting of 34 datasets
\end{enumerate}

\section{Related Works}

Gradient boosting decision trees (GBDT) have been the de facto method for tabular classification and regression tasks where XGBoost \cite{chen2016xgboost} dominates \cite{borisov2022deep}. This can mainly be attributed to the gradient boosting technique at scale as well as effective non-linear inductive biases. Yet, tasks involving textual data or a variable number of columns require specialized handling. NODE \cite{popov2019neural} offers a differentiable tree-like approach that outperforms shallow GBDTs but is significantly slower than gradient boosted trees. With the recent trends to leverage advances with the transformer architecture or pretrained language models, neural architectures become a popular choice.

TabNet \cite{arik2021tabnet}, motivated by the success of the transformer architecture \cite{vaswani2017attention}, uses an encoder-decoder architecture that can emulate the workings of decision trees and induce interpretability. TabTransformer \cite{huang2020tabtransformer} also uses a transformer with two separate input channels for categorical and continuous features, showing marginal gains over multi-layer perceptrons (MLP) on tabular classification and regression tasks. TabNet and TabTransformer work well but have to be tailored to one specific task at a time.

FT-Transformer \cite{gorishniy2021revisiting} followed a BERT-like \cite{devlin2019bert} approach with a classification token and a more elaborate tokenization strategy. SAINT \cite{somepalli2022saint} has a similar BERT-like design but extends with inter-sample attention and contrastive pretraining. However, both are unable to handle free-formed text and require latent embeddings per categorical feature.

TabPFN \cite{hollmann2022tabpfn} solves the more difficult task of in-context learning for tabular predictions. They adopted a Bayesian perspective and pretrained on synthetic datasets, showing success in time-constrained settings despite being unable to fully handle free-formed text. Two works on tabular data and time series are TabBert \cite{padhi2021tabular} and UniTTab \cite{luetto2023one} where temporal dependencies are modeled with a sequence transformer, but UniTTab extended with a sine-cosine numeric encoding and improved handling heterogeneous types. These models cannot handle values outside the predefined set of categories.

CARTE \cite{kim2024carte} views tabular data as graphlets. With graph attention networks and vast amounts of data for knowledge graphs, the pretrained model performs well in regression and classification tasks upon finetuning on small amounts of data. Nevertheless, it is reliant on its use of power transforms which limits its ability in multi-task domains. TabLLM \cite{hegselmann2023tabllm} and Tabula-8B \cite{gardner2024large} view tabular data as a language subtask. By inserting tabular information into a natural language template, pretrained large language models (LLM) are finetuned on datasets and directly used for predictions in the token space. Tabula-8B took it one step further by pretraining a Llama 3-8B model \cite{grattafiori2024llama} on over 2.1 billion rows of tabular data, making it a foundation model for tabular tasks. As a natural language subtask, it requires a large number of parameters, high memory and computational resource consumption, and LLMs sometimes struggle with numbers \cite{feng2024numerical}. We summarize the mentioned works in terms of their fulfillment of the desiderata in Table \ref{tab:relatedworksdesiderata}.

\begin{table}
	\caption{Summary of related works and whether they fulfill each desideratum. A \checkmark means that the algorithm fulfills that desideratum, otherwise it does not.\\}
	\label{tab:relatedworksdesiderata}
	\centering
	\hspace*{-5pt}
	\begin{tabular}[H]{rcccccc}
		\textbf{Algorithm} & \ref{d1} & \ref{d2} & \ref{d3} & \ref{d4} & \ref{d5} & \ref{d6}\\
		\midrule
		XGBoost \cite{chen2016xgboost}, NODE \cite{popov2019neural} &  &  &  &  & \checkmark & \checkmark \\
		TabNet, TabBert, UniTTab, FT-Tr., SAINT \cite{arik2021tabnet, padhi2021tabular, luetto2023one, gorishniy2021revisiting, somepalli2022saint} &  & \checkmark &  & \checkmark & \checkmark &  \\
		TabPFN \cite{hollmann2022tabpfn}, TabTransformer \cite{huang2020tabtransformer} &  &  &  &  & \checkmark &  \\
		CARTE \cite{kim2024carte} & \checkmark & \checkmark & \checkmark & \checkmark & \checkmark &  \\
		TabLLM \cite{hegselmann2023tabllm} and Tabula-8B \cite{gardner2024large} & \checkmark & \checkmark & \checkmark &  &  & \checkmark \\
		Ours & \checkmark & \checkmark & \checkmark & \checkmark & \checkmark & \checkmark \\
		\bottomrule
	\end{tabular}
\end{table}

\section{Background}

\subsection{Sign-Magnitude Representation}\label{sec:smr}

\textit{Sign-magnitude representation} (SMR) is conceptualized in the area of computer hardware to represent real numbers \citep{goldberg1991every}. Let $v \in \R$ be a scalar value and $e_\text{num}$ be a sign-magnitude encoder.  Let $h$ be the number of \textit{high bits} and $\ell$ be the number of \textit{low bits}. High bits refer to the coefficients of natural number base-2 exponents, while low bits refer to the coefficients of negative base-2 exponents. We call $a_0$ the \textit{sign bit}. If $v$ can be expressed in terms of $(-1)^{a_0} \cdot (a_1 2^{h-1} + a_2 2^{h-2} + \cdots + a_{h} 2^0 + a_{h+1} 2^{-1} + a_{h+2} 2^{-2} + \cdots + a_{h+\ell} 2^{-\ell})$ for some coefficients $a_0, a_1, ..., a_{h+\ell} \in \{0, 1\}$, then the encoding of $v$ is $e_\text{num}(v) = \begin{bmatrix}a_0 & a_1 & \cdots & a_{h+\ell}\end{bmatrix}$.


\subsection{Induced Set Attention}\label{sec:isa}

The induced set attention block is introduced by \citet{lee2019set} to overcome the quadratic time complexity in self-attention on input $X \in \R^{n \times d}$. They defined $m$ inducing points $I \in \R^{m \times d}$ used as a query in a cross attention layer and yielding $H \in \R^{m \times d}$. Finally, another cross attention layer is used where $X$ is the query and $H$ is the key-value, producing $O$ as the final output.
\begin{align}
	H = \text{CrossAttn}(I, X) \in \R^{m \times d} && O = \text{CrossAttn}(X, H) \in \R^{n \times d}
\end{align}
where two arguments to CrossAttn are the query and key-value respectively. In our works, we are only concerned with $H$, the one that produces fixed-size sequences.

\section{Methodology}

\subsection{Inputs}\label{sec:structure}


We view a row as an unordered set of column name $c_j$ and entry value $v_j$ pairs $\{(c_j, v_j)\}_{j=1}^C$, where $C$ is the number of columns. The set formulation directly fulfills \ref{d4}, and any model that treats it as a variable length sequence can handle a row from any table, fulfilling \ref{d2}. The $(c_j, v_j)$ pairs fulfill \ref{d3}.

We process all column names $c$ and entry values $v$ separately but in the same fashion. We encode numeric values using sign-magnitude representation (SMR) and textual inputs using BERT \citep{devlin2019bert} which are then passed into the model, fulfilling \ref{d1}. We use two linear projection layers to downscale BERT vectors and upscale the numeric vectors to embed them onto the same representation space. This also helps to differentiate between text and numbers.

If all columns are of numeric types then $L = 1$ since SMR is a single vector. Otherwise, $L$ is set to the maximum length and the entries are padded accordingly. Every missing value (NaN) is replaced with a special learnable token of length 1. Therefore, the dimensions of the encoded column names $c$ and encoded values $v$ of a row are:
\begin{align}
	e_\text{input}(c) \in \R^{B \times C \times L \times D} && e_\text{input}(v) \in \R^{B \times C \times L \times D}\label{eqn:rowencoding}
\end{align}
where $B$ refers to the batch size, $C$ refers to the number of columns, $L$ refers to the maximum sequence length of the entry values, and $D$ refers to the embedding dimension of each token.


\subsection{Architecture}

\begin{figure}
	\centering
	\hspace*{-13pt}
	\includegraphics[width=1.05\textwidth]{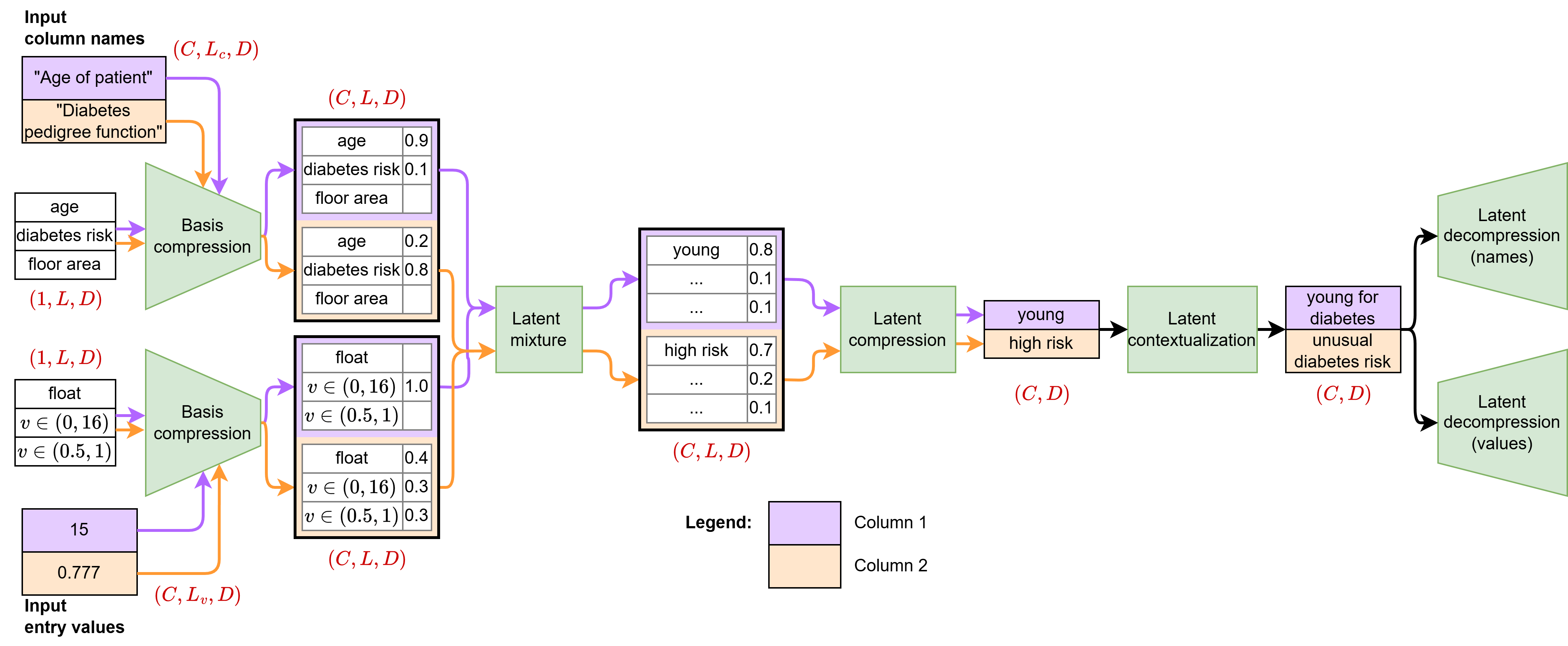}
	\caption{A basis transformer block, and a simplistic two-column diabetes example annotated with human interpretation to illustrate the flow of information. All texts except the two inputs are our interpretation. The general dimensions of vectors or sequences are denoted in red. The green blocks are the modules in a \BT block; the two leftmost white boxes are the learned basis vectors; the other white boxes denote vectors in a sequence and their respective mixture proportions; coloured arrows refer to independent operations by column; black arrows refer to inter-column operations.}
	\label{fig:combinedexample}
\end{figure}

We call our proposed architecture: basis transformers (\BT). The inputs to \BT are the two embeddings in Equation \ref{eqn:rowencoding}. The output is the SMR of a numeric target. In a top-down manner, we introduce the architecture followed by the details of the novel components in subsequent subsections. A \BT is comprised of stacking multiple \BT blocks sequentially, shown in Figure \ref{fig:combinedexample}.

\paragraph{Basis compression} \textit{Basis queries} are equivalent to inducing points in Section \ref{sec:isa}, but with different interpretations. They are used to summarize text and numeric values into a fixed-size sequence in terms of the learned bases. The first \BT block has a set of learnable basis queries of size $\R^{L \times D}$ where it will be broadcasted to $\R^{C \times L \times D}$ so that every column receives the same basis queries. Similar to the first cross attention layer in Section \ref{sec:isa} (the one producing $H$), \textit{basis compression} is a cross attention with basis queries as the query and input sequence as the key-value for each column independently.

\paragraph{Latent mixture} This module merges information between compressed column name-entry value pairs $(c_j, v_j)$ using cross attention. A number on its own means nothing. On the other hand, the interpretation of ``15'' under the ``age'' column could be ``young''. This is an intra-column operation, i.e. the cross attention is done between the column name and entry value of a single column. Information is processed in accordance with the intended tabular structure, satisfying \ref{d5}. Furthermore, the output is a set of encoded $(c_j, v_j)$ pairs, following the formulation in Section \ref{sec:structure}.

\paragraph{Latent compression} This module compresses a sequence of vectors into a single vector by concatenating the vectors then projecting to a lower dimensional subspace, so that the next module can operate on a high-level representational space.

\paragraph{Latent contextualization} This module processes inter-column relationships to extract high-level details needed for the task using self-attention layers. The number of self-attention layers is a hyperparameter dependent on the complexity of the task, similar to depth in decision trees.

\paragraph{Latent decompression} This module produces two new sets of basis queries of dimension $\R^{C \times L \times D}$ for the next block, enabling the chaining of multiple blocks sequentially. A table can have many columns, so the model has to ``look'' at the input row multiple times, each time viewing from a different perspective. The results of an ablation study on the number of \BT blocks can be found in Appendix \ref{sec:ablationblocks}. The final \BT block will only contain one latent decompression module, yielding a sequence of size $\R^{C \times L \times D}$ where mean aggregation will be performed over columns to get a tensor of size $\R^{L \times D}$. Finally, it is flattened and downscaled to a vector of size $D$ representing the prediction.

\subsection{Numeric Encoding}\label{sec:numencoding}


\paragraph{Motivation} Recall \ref{d1}. Numeric values in a column are typically scalars while textual values are encoded as high-dimensional vectors. We want to represent numbers as vectors so that a model can learn an effective joint representational space of numeric and textual data. Other works have treated numbers as strings \cite{hegselmann2023tabllm} or mapped them to learned latent transformation \cite{somepalli2022saint}. We use SMR as described in Section \ref{sec:smr}. An ablation using SMR and other encoding schemes on a simple task can be found in Appendix \ref{sec:ablationnumeric}.

\paragraph{Resolution} This representation splits a scalar into a vector of bits where each non-sign bit specializes on a certain magnitude scale which we call \textit{resolution}. The last bit focuses on small values while the second bit focuses on large values. Resolution-aware property can be useful in cases where \ref{d6} is needed. This also works well with attention, e.g. to select positive integer values, a query vector would be one where the last $\ell$ bits are $-1$ while the rest of the bits are $+1$.

\paragraph{Training} SMR is used for inputs and outputs. The output being a vector of binary values transforms the regression problem into a multi-label classification problem, hence we use the sum of binary cross entropy losses across all the $1 + h + \ell$ dimensions of the prediction. Hence, we do not need any form of normalization or scaling, satisfying \ref{d6} as well as potentially unlocking zero-shot capabilities.

An implication of this is each resolution is weighted equally in the loss function, resulting in a stable training process whereas mean squared error loss could grow intractably large and is sensitive to different scales in values. Targets likely differ in ranges of values in multi-task settings. For example, datasets with targets in $[10^{0}, 10^{1}]$ would be systematically underfitted compared to datasets with targets in $[10^{3}, 10^{6}]$ when using conventional regression losses. To see this phenomenon, an ablation study using MSE and BCE losses is conducted in Appendix \ref{sec:ablationscalar}.

To the best of our knowledge, transformation from regression tasks to multi-label classification tasks has never been done. There are works that convert a regression problem into a multi-class classification problem by predicting the interval containing the target \cite{lathuiliere2019comprehensive} but this is an approximation.

\subsection{Compression for Structured Data}\label{sec:basisqueries}



\paragraph{Objective} \ref{d1} and \ref{d2} call for a model that can natively handle heterogeneous types and variable columns. We want a method that can convert them to uniform structures. Secondly, we want an efficient modeling mechanism. Thirdly, we might want to convert structured data into a lower dimensional space, either for more efficient computation or to extract high-level representation, but also respecting the structure in reference to \ref{d5}.

\paragraph{Basis compression} Basis compression is done in linear time with respect to the input sequence length. It homogenizes columns into fixed-sized sequences, making it easier to handle. Upon gradient-based optimization, we conjecture the set of basis queries $q$ orients towards the most important parts of the embedding space of all input sequences. The choice of $q$ influences the degree of attention towards certain keywords, values, or a mixture of both. This emulates a ``soft" split in decision trees. For example, when predicting house prices from free-formed text, $q$ could contain vectors such as ``beautiful" and ``pests". This ensures that words of higher mutual information with the targets get a higher mixture proportion in the result.

\paragraph{Latent compression} To achieve the third objective, we have to understand the properties of the output of basis compression. Cross attention is an order equivariant operation with respect to the query and an order invariant operation with respect to the key-value (i.e. input sequence). Also, the order of basis queries is fixed by construction. Hence, the output of basis compression is always invariant to the order of the input sequence. Latent compression exploits this property by concatenating the output vectors of basis compression in the sequence then projecting it to a lower dimensional subspace.


\subsection{Module Implementation}\label{sec:moduledetails}



This section outlines the implementation details of the architecture. An \textit{independent dimension} is one where there is no information exchange between elements across that dimension, and operations are done in parallel e.g., batch size is an independent dimension.

\paragraph{Basis Compression} This module is a pre-layer normalization transformer layer specified by \citet{xiong2020layer} without self-attention layers, and using basis queries $q_x \in \R^{C \times L \times D}$ in the cross attention layer. For input sequence (column name or entry value) $X \in \R^{C \times L_x \times D}$,
\begin{align*}
	z_x = \texttt{BasisComp}(q_x, X) \in \R^{C \times L \times D}
\end{align*}

The cross attention is done over the $L_x$ dimension. $C$ is an independent dimension. If this is the first \BT block, $q_x \in \R^{L \times D}$ is duplicated (broadcasted) in the first dimension from 1 to $C$.

\paragraph{Latent mixture} This module consists of cross attention (over the $L$ dimension), self-attention (over the $L$ dimension) and layer normalizations. In the cross attention layer, the compressed column embedding is the query while the compressed value embedding is the key-value of each column. Then, the output goes through a self-attention layer for an improved mixture. $C$ is an independent dimension. For some compressed column embedding $z_\text{col} \in \R^{C \times L \times D}$ and compressed value embedding $z_\text{val} \in \R^{C \times L \times D}$,
\begin{align*}
	z = \texttt{LMix}(z_\text{col}, z_\text{val}) \in \R^{C \times L \times D}
\end{align*}

\paragraph{Latent compression} This module involves reshaping ($\R^{C \times L \times D}$ to $\R^{C \times LD}$) and a linear projection ($\R^{C \times LD}$ to $\R^{C \times rD}$) where $r \in \N$ is a hyperparameter. $C$ is an independent dimension.
\begin{align*}
	\bar{z} = \texttt{LComp}(z) \in \R^{C \times rD}
\end{align*}

\paragraph{Latent contextualization} To promote the learning of relationships between columns, multiple self-attention blocks (a self-attention layer over the $C$ dimension, a shallow MLP, and layer normalizations) are stacked sequentially. This module does not change dimensions.
\begin{align*}
	\bar{z}_\text{ctx} = \texttt{LCont}(\bar{z}) \in \R^{C \times rD}
\end{align*}

\paragraph{Latent decompression} The purpose of this module is to generate two new sets of basis queries for the next block. The intermediate latent tensor is first upscaled (from $\R^{C \times rD}$ to $\R^{C \times LD}$) with a linear projection layer then reshaped to $\R^{C \times L \times D}$. Finally, to specialize the tensor into name and value entities, it is passed through two identical but distinct sets of MLP and layer normalization.
\begin{align*}
	q_\text{col}, q_\text{val} = \texttt{LDecomp}({\bar{z}'}_\text{ctx}) \in \R^{2 \times C \times L \times D}
\end{align*}

\subsection{Adaptive Loss Reweighing}

In a multi-task regime where some tasks are inherently more challenging than others, we designed an \textit{adaptive loss reweighing scheme} to focus on hard examples for regression tasks during mini-batch training. For prediction $\hat{y} \in \R$ (decoded from SMR) and target $y \in \R$, we compute a bounded heuristic $g$ on how well the magnitude of prediction is
\begin{align}
	g(y, \hat{y}) = \frac{\min\{|\hat{y}|, |y|\} + \varepsilon}{\max\{|\hat{y}|, |y|\} + \varepsilon} \in (0, 1] \qquad \text{for a small } \varepsilon > 0
\end{align}

A hard example would have $g(y, \hat{y})$ close to 0. The new loss $\tilde{L}(y, \hat{y})$ over a mini-batch of $n$ samples based on the original loss function $L(y, \hat{y})$ is
\begin{align}\label{eqn:reweight}
	\tilde{L}(y_i, \hat{y}_i) & = \left[(1 - g(y_i, \hat{y}_i)) \cdot (1 - 2\gamma) + \gamma\right] L(y_i, \hat{y}_i)
\end{align}

where $\gamma \in [0, 0.5]$ controls the aggressiveness of loss reweighing by rescaling $1 - g(y, \hat{y})$ to a range of $[\gamma, 1-\gamma)$. While this scheme ignores the sign of a prediction, in practice, most models can predict the signs well especially \BT which has a dedicated bit for sign prediction. Appendix \ref{sec:adaptiveloss} contains a more detailed discussion of the scheme. An ablation study of various $\gamma$ is done in Appendix \ref{sec:ablationadaptiveloss}.

\section{Experiments}

\subsection{Experimental Details}


The experiments use the OpenML-CTR23 benchmark \cite{fischer2023openml}, consisting of 34 regression datasets (one dataset was removed due to the large number of columns; see Appendix \ref{sec:hyperparams} for more details). The entry values consist of numbers, textual categorical values, and free-formed text. No standard preprocessing is performed, such as one-hot encoding, standard scaling of mean and standard deviation, and dropping rows with not-a-numbers (NaNs). Each dataset is split into training, validation, and test sets. 5 random seeds are used for each experiment. For each seed, the model evaluated on the test set is selected using the best mean score based on the validation set.

The primary evaluation metric is the coefficient of determination, commonly denoted as $R^2$. The range of values is $(-\infty, 1]$. There are three regions/points in $R^2$ of interest \cite{hahn1973coefficient}: 1 means the model perfectly explains the variance, 0 means the model explains none of the variance but has the correct mean prediction, and a negative score means being worse than a naive mean-based predictor. $R^2$ offers a scaled measure of performance across different ranges of targets, leading to better interpretation of results, unlike mean squared error which is not invariant to scale transformations. Each dataset will have its own $R^2$ score. See Appendix \ref{sec:completeexp1} for further details about how the $R^2$ statistics are computed.

\subsection{Multi-Task Experiments}


A model is trained on the entire OpenML dataset simultaneously. After every 200 gradient steps, the model is validated on each dataset. Finally, the best-performing checkpoint is tested on each dataset.

The baselines used to evaluate against \BT are large language models as prescribed by TabLLM \cite{hegselmann2023tabllm}. We use the best-performing serialization technique they achieved, which is the template \texttt{The <COLUMN\_NAME> is <VALUE>.}. The LLMs are finetuned on rows that are serialized with the template. For a fair comparison, we limit the number of parameters of baselines up to 5 times the size of \BT. There are two classes of LLMs that we use as baselines: sequence-to-sequence language models (Flan-T5 \cite{chung2024scaling}, LaMini-T5 \cite{lamini-lm}, BART-large \cite{lewis2020bart}), and causal language models (Pythia \cite{biderman2023pythia}, Cerebras-GPT \cite{dey2023cerebras}). We also considered comparisons to Qwen, TinyLlama, TABULA-8B and CARTE, but we had to exclude them due to data contamination issues (see Appendix \ref{sec:invalidbaselines}).  We also had to exclude XGBoost \cite{chen2016xgboost}, TabNet \cite{arik2021tabnet}, NODE \cite{popov2019neural}, TabPFN \cite{hollmann2022tabpfn}, TabTransformer \cite{huang2020tabtransformer}, TabKANet \cite{gao2024tabkanet}, TabBert \cite{padhi2021tabular}, UniTTab \cite{luetto2023one}, UniPredict \cite{wang2023unipredict}, FT-Transformer  \cite{gorishniy2021revisiting}, SAINT \cite{somepalli2022saint} and Tabular-Text Transformer \cite{bonnier2024revisiting} because they either could not handle free form text, could not deal with heterogeneous rows, required specialized information or were not reproducible (see Appendix \ref{sec:invalidbaselines}).

The weights of \BT were randomly initialized while the LLMs were pretrained on their respective pretraining tasks, and the checkpoints were obtained from Huggingface \cite{wolf2020huggingface}. Hyperparameters and configurations are thoroughly described in Appendix \ref{sec:hyperparams}.

\begin{table}
	\caption{Central tendency and spread statistics of test $R^2$ scores across 34 datasets. Higher median and mean are better; lower IQR and standard deviation are better.\\}
	\label{tab:exp1stats}
	\centering
	\begin{tabular}{lllll}
		& Median & Interquartile range & Mean & Standard deviation \\
		\midrule
		Basis Transformer (Ours) & $\bm{0.241}$ & $0.989$ & $\bm{\text{-}0.565}$ & $\bm{3.295}$ \\
		Pythia410M & $\text{-}2.980$ & $23.490$ & $\text{-}160.334$ & $528.735$ \\
		Pythia1B & $\text{-}7.892$ & $53.624$ & $\text{-}8.96 \times 10^{30}$ & $5.14 \times 10^{31}$ \\
		Cerebras590M & $\text{-}0.162$ & $3.564$ & $\text{-}6.326$ & $12.883$ \\
		Cerebras1.3B & $\text{-}0.097$ & $1.982$ & $\text{-}123.694$ & $566.072$ \\
		FlanT5-base & $\text{-}0.280$ & $1.191$ & $\text{-}2.204$ & $4.941$ \\
		LaMiniT5-223M & $\text{-}0.157$ & $\bm{0.471}$ & $\text{-}2.663$ & $8.694$ \\
		BART-large & $\text{-}1.697$ & $10.784$ & $\text{-}4.01 \times 10^{9}$ & $2.31 \times 10^{10}$ \\
		\bottomrule
	\end{tabular}
\end{table}

Table \ref{tab:exp1stats} reports central tendency and spread, as well as robust and non-robust statistics. \BT outperforms in central tendency in both robust and non-robust measures. It also shows \BT and the T5 derivatives are relatively precise in a way that their performances do not vary as much. In contrast, Cerebras1.3B has the second-best median but incredibly poor mean performance due to performing badly on certain datasets. To view the complete breakdown of results by datasets, standard deviation over random seeds, and runtimes, they are shown in Appendix \ref{sec:completeexp1}. The hardware used is reported in Appendix \ref{sec:hardware}.

\begin{figure}
	\centering
	\includegraphics[width=0.9\textwidth]{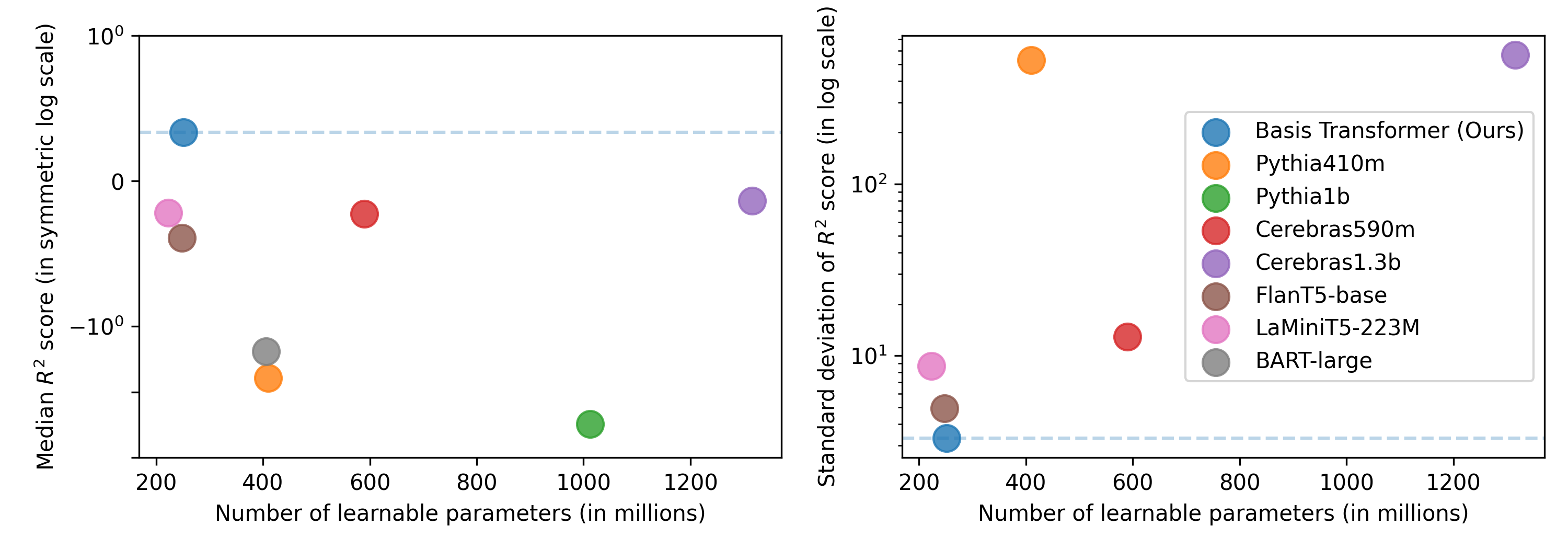}
	\caption{Median and standard deviation of $R^2$ against the number of learnable parameters. High standard deviation ($> 10^8$) results are omitted as outliers.}
	\label{fig:r2params}
\end{figure}

To give a sense of performance as a function of the number of learnable parameters, Figure \ref{fig:r2params} shows that \BT has the highest median $R^2$ score despite being one of the smallest models and without any pretraining. Furthermore, the right side of the figure shows a trend in how larger models tend to have much higher variability in performance across datasets.

\begin{table}
	\caption{Mean and standard deviation of success rate over 5 random seeds across datasets.\\}
	\label{tab:exp1success}
	\centering
	\begin{tabular}{lllllllll}
		& Ours & Py410m & Py1b & Ce590m & Ce1.3b & FlanT5 & LMT5 & BART \\
		\midrule
		Mean & $\bm{1.00}$ & $0.95$ & $0.89$ & $0.92$ & $0.90$ & $\bm{1.00}$ & $\bm{1.00}$ & $0.98$ \\
		Standard deviation & $\bm{0.00}$ & $0.12$ & $0.27$ & $0.21$ & $0.19$ & $\bm{0.00}$ & $\bm{0.00}$ & $0.06$ \\
		\bottomrule
	\end{tabular}
\end{table}

The LLMs do not always abide by the serialization template. During evaluation, the generated outputs by the LLMs are parsed using a lenient regular expression to extract the numeric values, ignoring: excessive white spaces, lack of an EOS token, or a missing period. The proportion of predictions that can be parsed is called \textit{success rate}, which is reported in Table \ref{tab:exp1success}. In general, sequence-to-sequence models do better than causal models.

\begin{figure}
	\centering
	\hspace*{-13pt}
	\includegraphics[width=1.05\textwidth]{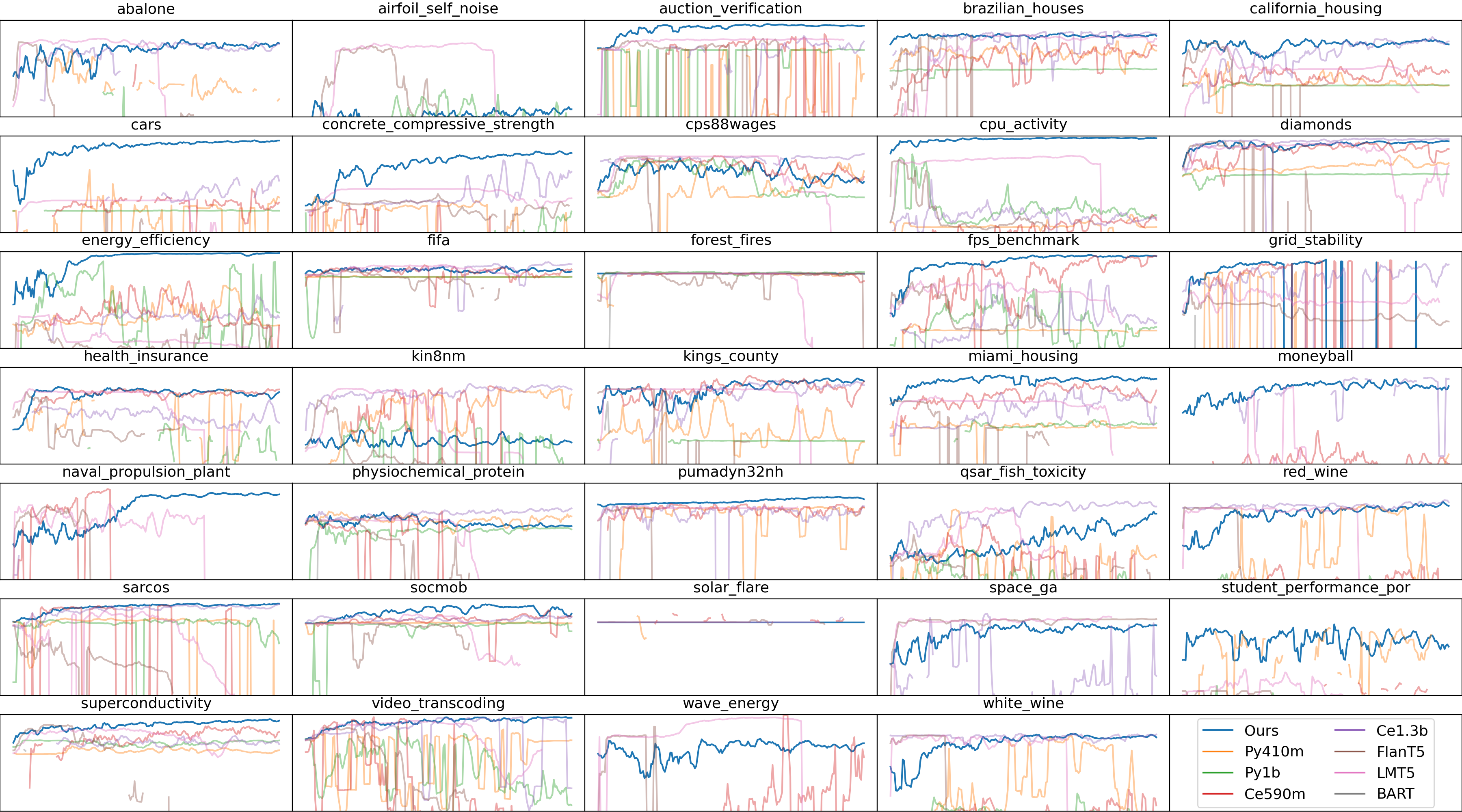}
	\caption{Validation $R^2$ scores over gradient steps per dataset. The range of $y$-axis is [-20, 1] in symmetric log scale and the range of $x$-axis is [0, 200]. The blue line is \BT. Higher values are better.}
	\label{fig:evalsteps}
\end{figure}

Another important point is the stability of training, shown in Figure \ref{fig:evalsteps}. After every 200 gradient steps, each model is evaluated once on all validation datasets. There are 40,000 gradient steps for each model. Some lines from the baselines are discontinuous because they fail to conform to the prescribed template at some points, hence they cannot be evaluated. \BT is consistently stable when evaluated on unseen samples. Despite being trained from scratch, it acquired a decent understanding of the tasks within 200 gradient steps (i.e. the left-most blue point in every subplot), likely due to the numeric representation already being determined. For an enlarged landscape view and information on variability across random seeds, Figure \ref{fig:evalstepswithspread} can be found in the Appendix \ref{sec:completeexp1}.





\subsection{Insights}


\paragraph{Memory consumption}

A limitation of \BT is the memory consumption, attributed to its 4D input tensor representation. As transformers require padding for parallelism during mini-batch training, a higher dimension tensor leads to more wastage. This is especially the case where the variance in sequence length $L$ is high as \BT handles numeric values ($L = 1$), and textual values ($L \gg 1$).

\paragraph{Alignment to multi-task Bayesian related works}

Multi-task tabular regression is challenging because a model has to: (1) infer the task from a single row, (2) digest information from the row given the task, and (3) calibrate the prediction onto an unbounded range of values given input information and task. The Bayesian perspective of multi-task learning with a similar process has been explored \cite{bishop2012bayesian, wilson2007multi, swersky2013multi} primarily on Gaussian processes, but the architecture of \BT abides by some key premises. It has been noted that not all tasks are equally similar in practice \cite{bakker2003task, daume2009bayesian}; there are inherent structures within certain groups of tasks. Basis queries align with this where a task is implicitly defined by the set of column names and entry values, whereupon decomposed into a sequence, can be expressed in terms of common identifiers potentially across all tasks.

\paragraph{The future of foundation model training}

Figure \ref{fig:evalsteps} shows the fluctuations in the ability of LLMs to generalize in tabular tasks. The entire premise of Tabula-8B \cite{gardner2024large} is making a foundation model for tabular tasks using an 8 billion parameter Llama 3 model \cite{grattafiori2024llama}, and is mainly evaluated at its test time performance in the paper. One should question how effective the training is in establishing good learned representation when pretraining. Pretraining a foundation model is essentially multi-task learning but at a much larger scale. In Figure \ref{fig:r2params}, we see that the standard deviation of performance grows as the model size increases, likely attributed to a poor choice of numeric representation.


\section{Conclusion}


The design desiderata, numeric representation, architecture, and training routine build up to a model that outperforms LLMs in multi-task tabular regression. Besides raw performance, the lightweight design, efficient forward pass, and stable training are non-trivial achievements. While already showing success in 34 tasks, we believe there is great potential in fully scaling up to a foundation model with these principles set in place. See Appendix \ref{sec:societalimpacts} for a discussion of societal impacts.

\newpage

\begin{ack}
Resources used in preparing this research were provided, in part, by the Province of Ontario, the Government of Canada through CIFAR, and companies sponsoring the Vector Institute \url{http://www.vectorinstitute.ai/\#partners}.
\end{ack}

\bibliography{bibliography}


\newpage

\appendix

\section{Discussion of Societal Impacts}\label{sec:societalimpacts}

While the scope of the paper focuses narrowly on tabular modality, the broader impact could be extended to both good and bad to humankind. Tabular data is the standard in commercial and industrial applications, which contains extensive personal information. Recall that our works in this paper ultimately aim to set key design principles in models for tabular data, where if successful, might lead to an overall interest in tabular models and hence greater demand for tabular data. This could lead to negatives if the information could be used for illegitimate surveillance of the general population or other malicious activities. However, personal information such as medical records or economic/finance data that are currently left underutilized could finally be leveraged for the betterment of groups of people in need.

\section{Ablation on Numeric Encoding Schemes}\label{sec:ablationnumeric}

Prior to the full experiments, we believe it is important to validate our hypothesis that sign-magnitude representation is superior over the IEEE 754 32-bit floating point representation and latent embeddings learned from a neural network.

For each random seed, 1000 random numbers are generated satisfying two conditions. Each number is between -100 and 100. Each number can be described as at least one of: even, odd, real, integer, big (absolute value more than 50), or small (absolute value less than or equal to 50). This is captured with a multi-hot vector for each number.

This is a multi-label classification task where a model maps a representation of a number to the multi-hot target. For consistency, sign-magnitude and IEEE 754 representations (both 32 bits) use a 2 hidden layer multi-layer perceptron. The latent embedding method uses a 3 hidden layer multi-layer perceptron with the scalar directly fed in so the first hidden layer essentially produces an embedding. All models are trained for 250 epochs and are validated with 1000 random numbers using the cross entropy loss as a metric.

\begin{figure}[H]
	\centering
	\includegraphics[width=0.48\textwidth]{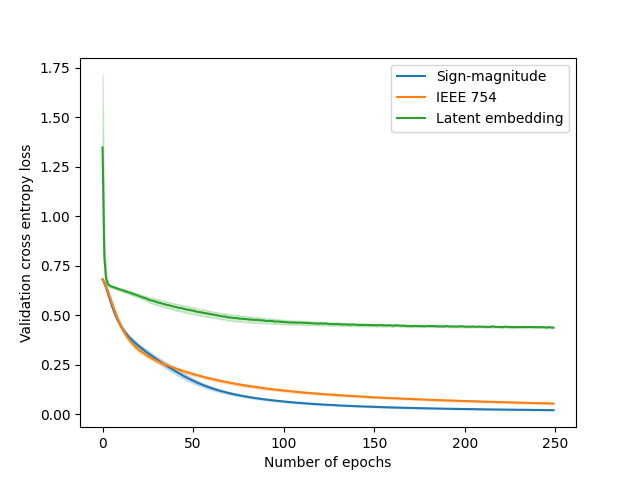}
	\caption{Relative performance (with a standard error of $1.96\sigma$) of different numeric encoding schemes on the same task over 10 random seeds. The models for sign-magnitude and IEEE 754 are initialized with the same random weights for each seed.}
	\label{fig:numericencoding}
\end{figure}

Figure \ref{fig:numericencoding} demonstrates that sign-magnitude consistently learns faster than the others. While this is a simple problem and the gains are marginal over the IEEE 754 representation, we believe this improvement is more prominent in more complex problems using numeric values.

\section{Rationale behind Adaptive Loss Reweighing Scheme}\label{sec:adaptiveloss}

There are several regression metrics that could be used to reweigh the loss such as $R^2$, normalized Nash-Sutcliffe Efficiency (NNSE) \cite{gupta2011typical} and Robinson's Agreement Coefficient (RAC) \cite{robinson1957statistical}. These regression metrics are rescaled to a bounded interval using a measure of variance. Using variance to standardize the metric is not suitable for a mini-batch process, especially in a multi-task learning regime. The variance of targets wildly fluctuates across tasks, leading to inaccurate metrics in a mini-batch.

Our approach measures performance as a ratio of prediction to target or a ratio of target to prediction, whichever is smaller. This ensures that the measure is bounded to $[0, 1]$, and does not have a dependency on statistics of other elements in the mini-batch.

\section{Invalid Baselines and Justifications}\label{sec:invalidbaselines}

Given the scope of our paper, there are several possible baseline algorithms that fail to meet one or more criteria. We will first discuss the two special cases, followed by the general cases that most other works fall under.


\subsection{Special Cases}

Qwen2.5 \cite{yang2024qwen2} is a fairly recent LLM that is competitive with many state-of-the-art LLMs. Our experiments with Qwen showed an unprecedented superior performance -- some datasets reached above 0.8 validation $R^2$ score when trained on less than 8\% of training data. After training on 17\% of the samples, it has almost perfect predictions (close to 0 mean squared error) on 5 tasks (validation set). Upon inspection, their technical report mentioned that they trained with tables as part of their efforts to improve support to a larger domain. This confirms our suspicion that Qwen, unfortunately, could not be used as a baseline algorithm due to data contamination. This is also the case for TinyLlama \cite{zhang2024tinyllama} which was pretrained on The Stack dataset \cite{kocetkov2022stack} which uses CSV data and attained impossibly high validation score upon being finetuned on small amounts of training data. It attained perfect predictions on one task (validation set) after training on 3\% of the samples. By 12\% of the training samples, it has attained perfect predictions on 6 tasks (validation set).

TABULA-8B \citep{gardner2024large} follows a very similar direction as TabLLM \citep{hegselmann2023tabllm}. The primary difference is TABULA-8B was pretrained on over 2.1 billion rows of tabular data \citep{gardner2024large}, while the models in TabLLM were only pretrained on language tasks. The OpenML CTR-23 dataset \cite{fischer2023openml} is a very popular and common tabular dataset. In Table 1 of \citet{gardner2024large}'s paper, it specifically mentions that 34.2\% of OpenML-CTR23 datasets has been potentially leaked. To avoid data contamination, this is the main reason TABULA-8B was not included as a baseline algorithm.

CARTE \cite{kim2024carte} has a novel design that views a row of a table as a graphlet. We were eager to compare our works with CARTE, but we faced serious implementation difficulties when applying it to a multi-task regression setting as a result of their design. Unlike other works that use LLMs for tabular tasks, numeric values are represented by taking the embedding of its corresponding column name embedding and scaling it by the scalar value. In the paper, they mentioned that numeric values are preprocessed beforehand using a power transform. The multi-task experiments require a model to be fitted with multiple datasets simultaneously, implying inconsistent ranges of target values. In addition to power transform not being a suitable preprocessing step, we faced floating point overflow issues when running multi-task experiments with a wide range of targets on the CARTE code base. When removing the power transformation preprocessing altogether, the training process begins but is plagued with NaNs due to very large loss values from datasets such as wave\_energy. The issue is not rectified even when dividing the value of all targets by a large constant. At that point, we decided not to go forward with more invasive modifications of their works at the risk of overriding the core of their works.

\subsection{General Cases}


There are four categories of design deficits in terms of requirements for the multi-task experiments.
\begin{enumerate}[label=(\Alph*)]
	\item Inability to handle free-form textual data in entries
	\item Inability to handle heterogeneous rows (inconsistent columns)
	\item Requires additional specialized information for each dataset
	\item No complete, reproducible code
\end{enumerate}

(A) is a common issue with many tabular models as many assume tabular data are mostly numeric. The OpenML CTR-23 dataset \cite{fischer2023openml}, however, contains tables that have more than 50\% textual columns. (B) This is a key requirement for multi-task learning. Many models are adapted to handle only fixed-size inputs, effectively ruling out being a multi-task or foundation model as they cannot generalize to multiple datasets. (C) While this is generally not an issue in tabular tasks, in the case of multi-task learning, this is highly restrictive. If a model requires auxiliary information such as the set of categorical features for certain columns, or a standard scaler for numeric values of certain columns, it diminishes the capacity of a multi-task or zero-shot model. (D) Without complete code provided by authors of baseline algorithms, it would be impossible to implement and experiment with the exact competitive configuration demonstrated by the authors.

The summary of invalid baseline algorithms is shown in Table \ref{tab:invalidbaselines}. Unlike Table \ref{tab:relatedworksdesiderata}, Table \ref{tab:invalidbaselines} is not about fulfilling desiderata but rather a concrete inability to implement for our experiments.


\begin{table}[H]
	\caption{Summary of invalid baseline algorithms and their justifications.\\}
	\label{tab:invalidbaselines}
	\centering
	\begin{tabular}[H]{rcccc}
		\textbf{Algorithm} & \textbf{(A)} & \textbf{(B)} & \textbf{(C)} & \textbf{(D)}\\
		\midrule
		Gradient boosting decision trees (e.g. XGBoost \cite{chen2016xgboost}) & \checkmark & \checkmark & & \\
		TabNet \cite{arik2021tabnet} & & \checkmark & \checkmark & \\
		NODE \cite{popov2019neural} & \checkmark & \checkmark & & \\
		TabPFN \cite{hollmann2022tabpfn} & \checkmark & \checkmark & &\\
		TabTransformer \cite{huang2020tabtransformer} & \checkmark & \checkmark & & \\
		TabKANet \cite{gao2024tabkanet} & \checkmark & \checkmark & \checkmark & \\
		TabBert \cite{padhi2021tabular} & \checkmark & & \checkmark & \\
		UniTTab \cite{luetto2023one} & \checkmark &  & \checkmark & \checkmark \\
		UniPredict \cite{wang2023unipredict} & & & & \checkmark \\
		FT-Transformer \cite{gorishniy2021revisiting} & \checkmark & & \checkmark & \\
		SAINT \cite{somepalli2022saint} & \checkmark & & \checkmark & \\
		Tabular-Text Transformer \cite{bonnier2024revisiting} & & & \checkmark & \\
		\bottomrule
	\end{tabular}
\end{table}

Given the circumstances, large language models are the only comparable baselines for our works.

\section{Complete Multi-Task Results}\label{sec:completeexp1}

\subsection{Computation of Scores}

Recall that all experiments are done over 5 random seeds. Each experiment consists of 34 datasets and 8 algorithms. All measures of $R^2$ scores are computed using the mean over random seeds. Central tendencies and spreads in Table \ref{tab:exp1stats} are effectively statistics of the mean $R^2$ scores. That means the reported statistics are in fact median of means, IQR of means, mean of means, and standard deviation of means. For brevity and clarity, we treat the mean over random seeds as the actual score.

However, in the following subsections, we expand to show the means and standard deviation over the random seeds.

\subsection{Scores by Dataset and Algorithm}

Abbreviations are used here so that the table can be formatted within the margins.
\begin{itemize}
	\item Py410m -- Pythia410M
	\item Py1b -- Pythia1B
	\item Ce590m -- Cerebras-GPT-590M
	\item Ce1.3b -- Cerebras-GPT-1.3B
	\item FlanT5 -- Flan-T5-base
	\item LMT5 -- LaMini-T5-223M
	\item BART -- BART-large
\end{itemize}

In Table \ref{tab:exp1meanr2}, for some algorithms and for some datasets, the performance exceeds a range that makes it difficult to include in a compacted table. As such, we denote mean $R^2$ score worse than -100 by {\color{orange} \texttt{Poor}}. Similarly, in Table \ref{tab:exp1stdevr2}, standard deviations of $R^2$ score that exceeds 100 are denoted by {\color{orange} \texttt{Poor}}. Both tables have the dataset names truncated to 8 characters.

\begin{table}[H]
	\caption{Mean $R^2$ score over 5 random seeds by dataset and algorithm.\\}
	\label{tab:exp1meanr2}
	\centering
	\begin{tabular}{lllllllll}
		& Ours & Py410m & Py1b & Ce590m & Ce1.3b & FlanT5 & LMT5 & BART \\
		\midrule
		abalone & $\text{-}0.26$ & $\text{-}0.18$ & $\text{-}50.36$ & $\text{-}3.74$ & $\text{-}0.61$ & $\text{-}0.38$ & $\bm{ \text{-}0.03 }$ & $\text{-}9.73$ \\
		airfoil  & $\text{-}18.81$ & {\color{orange} \texttt{Poor}} & $\text{-}15.82$ & $\text{-}4.73$ & {\color{orange} \texttt{Poor}} & $\text{-}0.88$ & $\bm{ \text{-}0.37 }$ & $\text{-}0.41$ \\
		auction  & $\bm{ 0.74 }$ & $\text{-}0.64$ & {\color{orange} \texttt{Poor}} & $0.26$ & $\text{-}0.09$ & $\text{-}0.32$ & $\text{-}0.08$ & $\text{-}0.54$ \\
		brazilia & $0.40$ & $\text{-}1.41$ & $\text{-}1.72$ & $\text{-}0.65$ & $\bm{ 0.49 }$ & $\text{-}0.16$ & $0.24$ & $\text{-}0.36$ \\
		californ & $\text{-}0.69$ & $\text{-}2.62$ & $\text{-}2.99$ & $\text{-}1.31$ & $\bm{ \text{-}0.08 }$ & $\text{-}0.14$ & $\text{-}1.28$ & $\text{-}7.38$ \\
		cars & $\bm{ 0.77 }$ & $\text{-}4.96$ & $\text{-}5.33$ & $\text{-}2.27$ & $\text{-}1.79$ & $\text{-}2.64$ & $\text{-}42.63$ & $\text{-}57.32$ \\
		concrete & $\bm{ \text{-}0.23 }$ & $\text{-}4.06$ & {\color{orange} \texttt{Poor}} & $\text{-}1.13$ & $\text{-}0.38$ & $\text{-}5.10$ & $\text{-}3.10$ & $\text{-}4.58$ \\
		cps88wag & $\text{-}0.71$ & $\text{-}1.95$ & $\text{-}0.25$ & $\bm{ 0.11 }$ & $\text{-}0.06$ & $\text{-}0.09$ & $\text{-}0.01$ & $\text{-}0.19$ \\
		cpu acti & $\bm{ 0.88 }$ & $\text{-}17.64$ & $\text{-}10.45$ & $\text{-}23.40$ & $\text{-}7.17$ & $\text{-}16.14$ & $0.02$ & $\text{-}21.49$ \\
		diamonds & $0.64$ & $\text{-}1.10$ & $\text{-}1.19$ & $0.32$ & $\bm{ 0.87 }$ & $0.73$ & $0.60$ & $\text{-}1.00$ \\
		energy e & $\bm{ 0.91 }$ & $\text{-}4.61$ & $\text{-}0.23$ & $\text{-}1.35$ & $\text{-}2.55$ & $\text{-}3.56$ & $\text{-}5.16$ & $\text{-}4.40$ \\
		fifa & $\bm{ 0.70 }$ & $\text{-}0.21$ & $\text{-}0.21$ & $0.15$ & $0.39$ & $0.01$ & $0.23$ & {\color{orange} \texttt{Poor}} \\
		forest f & $\text{-}0.16$ & $\text{-}0.48$ & $\bm{ \text{-}0.01 }$ & $\text{-}0.16$ & $\text{-}0.16$ & $\text{-}0.16$ & $\text{-}0.16$ & $\text{-}0.17$ \\
		fps benc & $\bm{ 0.86 }$ & $\text{-}6.21$ & $\text{-}5.20$ & $0.06$ & $\text{-}2.05$ & $\text{-}1.38$ & $\text{-}0.58$ & $\text{-}2.30$ \\
		grid sta & $0.47$ & $\text{-}96.64$ & {\color{orange} \texttt{Poor}} & $\bm{ 0.59 }$ & $0.28$ & $\text{-}0.70$ & $\text{-}0.12$ & $\text{-}1.68$ \\
		health i & $\text{-}0.32$ & $\text{-}3.34$ & $\text{-}51.48$ & $\bm{ \text{-}0.12 }$ & $\text{-}1.57$ & $\text{-}1.20$ & $\text{-}0.46$ & $\text{-}2.43$ \\
		kin8nm & $\text{-}3.62$ & $\text{-}1.22$ & $\text{-}2.29$ & $\text{-}0.17$ & $\bm{ \text{-}0.08 }$ & $\text{-}0.24$ & $\text{-}0.11$ & $\text{-}0.34$ \\
		kings co & $0.44$ & $\text{-}2.54$ & $\text{-}3.25$ & $\bm{ 0.65 }$ & $0.45$ & $\text{-}0.06$ & $\text{-}0.04$ & $\text{-}11.72$ \\
		miami ho & $\bm{ 0.33 }$ & $\text{-}1.96$ & $\text{-}1.93$ & $0.21$ & $0.15$ & $\text{-}0.34$ & $\text{-}0.28$ & $\text{-}96.60$ \\
		Moneybal & $0.20$ & $\text{-}60.14$ & $\text{-}56.02$ & $\text{-}2.13$ & $\bm{ 0.33 }$ & $\text{-}23.02$ & $\text{-}30.46$ & {\color{orange} \texttt{Poor}} \\
		naval pr & $0.55$ & {\color{orange} \texttt{Poor}} & {\color{orange} \texttt{Poor}} & $\bm{ 0.70 }$ & {\color{orange} \texttt{Poor}} & $\text{-}0.15$ & $\text{-}0.30$ & $\text{-}0.02$ \\
		physioch & $\text{-}1.18$ & $\text{-}1.06$ & $\text{-}1.49$ & $\bm{ \text{-}0.12 }$ & $\text{-}0.29$ & $\text{-}1.12$ & $\text{-}1.12$ & $\text{-}1.71$ \\
		pumadyn3 & $\bm{ 0.28 }$ & {\color{orange} \texttt{Poor}} & {\color{orange} \texttt{Poor}} & $\text{-}0.08$ & $\text{-}0.88$ & $\text{-}0.49$ & $\text{-}0.49$ & $\text{-}1.36$ \\
		QSAR fis & $\text{-}1.44$ & $\text{-}3.70$ & $\text{-}31.92$ & $\text{-}4.09$ & $\bm{ 0.14 }$ & $\text{-}4.50$ & $\text{-}0.13$ & $\text{-}9.76$ \\
		red wine & $\text{-}0.21$ & $\text{-}21.91$ & $\text{-}32.35$ & $\text{-}32.22$ & $\bm{ \text{-}0.08 }$ & $\text{-}0.22$ & $\text{-}0.22$ & $\text{-}1.01$ \\
		sarcos & $0.65$ & $\text{-}0.15$ & $\text{-}0.27$ & $\bm{ 0.80 }$ & $0.51$ & $\text{-}1.39$ & $0.18$ & $\text{-}0.38$ \\
		socmob & $\bm{ 0.54 }$ & $\text{-}0.16$ & $\text{-}0.15$ & $0.22$ & $0.08$ & $\text{-}0.18$ & $\text{-}0.12$ & $\text{-}0.22$ \\
		solar fl & $\text{-}0.16$ & $\text{-}0.38$ & {\color{orange} \texttt{Poor}} & $\text{-}0.18$ & $\bm{ 0.33 }$ & $\text{-}0.16$ & $\text{-}0.16$ & $\text{-}0.57$ \\
		space ga & $\text{-}0.91$ & $\text{-}30.62$ & {\color{orange} \texttt{Poor}} & $\text{-}30.67$ & $\text{-}2.47$ & $\bm{ \text{-}0.07 }$ & $\text{-}0.33$ & {\color{orange} \texttt{Poor}} \\
		student  & $\bm{ \text{-}0.24 }$ & $\text{-}51.77$ & $\text{-}34.92$ & $\text{-}31.81$ & $\text{-}1.88$ & $\text{-}11.37$ & $\text{-}5.00$ & $\text{-}13.01$ \\
		supercon & $0.60$ & $\text{-}0.91$ & $\text{-}0.50$ & $0.12$ & $\text{-}0.56$ & $\bm{ 0.62 }$ & $0.28$ & $\text{-}1.00$ \\
		video tr & $0.74$ & $\text{-}0.61$ & $0.15$ & $0.73$ & $\bm{ 0.91 }$ & $0.09$ & $0.41$ & $0.34$ \\
		wave ene & $\text{-}0.63$ & {\color{orange} \texttt{Poor}} & {\color{orange} \texttt{Poor}} & $\text{-}50.33$ & {\color{orange} \texttt{Poor}} & $\text{-}0.04$ & $\bm{ 0.38 }$ & {\color{orange} \texttt{Poor}} \\
		white wi & $\text{-}0.37$ & $\text{-}25.28$ & $\text{-}25.38$ & $\text{-}29.37$ & $\bm{ \text{-}0.10 }$ & $\text{-}0.18$ & $\text{-}0.16$ & $\text{-}1.51$ \\
		\midrule
		median & $\bm{0.24}$ & $\text{-}2.98$ & $\text{-}7.89$ & $\text{-}0.16$ & $\text{-}0.10$ & $\text{-}0.28$ & $\text{-}0.16$ & $\text{-}1.70$ \\
		\bottomrule
	\end{tabular}
\end{table}

\begin{table}[H]
	\caption{Standard deviation of $R^2$ score over 5 random seeds by dataset and algorithm.\\}
	\label{tab:exp1stdevr2}
	\centering
	\begin{tabular}{lllllllll}
		& Ours & Py410m & Py1b & Ce590m & Ce1.3b & FlanT5 & LMT5 & BART \\
		\midrule
		abalone & $0.09$ & $0.17$ & $39.53$ & $2.49$ & $0.38$ & $0.13$ & $\bm{ 0.08 }$ & $2.11$ \\
		airfoil  & $13.81$ & $33.18$ & $17.79$ & $4.42$ & $27.31$ & $0.28$ & $\bm{ 0.03 }$ & $0.24$ \\
		auction  & $0.12$ & $0.14$ & {\color{orange} \texttt{Poor}} & $0.09$ & $0.23$ & $0.16$ & $\bm{ 0.07 }$ & $0.13$ \\
		brazilia & $0.12$ & $0.34$ & $0.14$ & $0.22$ & $0.11$ & $0.12$ & $\bm{ 0.05 }$ & $0.11$ \\
		californ & $0.31$ & $0.34$ & $0.40$ & $0.61$ & $0.12$ & $\bm{ 0.05 }$ & $0.18$ & $1.47$ \\
		cars & $\bm{ 0.11 }$ & $1.46$ & $1.12$ & $0.42$ & $0.53$ & $0.53$ & $79.85$ & $45.03$ \\
		concrete & $0.26$ & $0.44$ & $58.56$ & $0.47$ & $\bm{ 0.20 }$ & $0.62$ & $0.47$ & $0.60$ \\
		cps88wag & $0.40$ & $0.42$ & $0.29$ & $0.02$ & $0.09$ & $0.10$ & $\bm{ 0.01 }$ & $0.18$ \\
		cpu acti & $\bm{ 0.07 }$ & $10.63$ & $5.77$ & $5.67$ & $2.87$ & $6.24$ & $0.07$ & $7.68$ \\
		diamonds & $0.29$ & $0.20$ & $0.12$ & $0.26$ & $\bm{ 0.03 }$ & $0.04$ & $0.06$ & $0.38$ \\
		energy e & $\bm{ 0.04 }$ & $0.51$ & $0.62$ & $0.18$ & $0.41$ & $0.38$ & $0.97$ & $0.72$ \\
		fifa & $0.06$ & $0.07$ & $0.07$ & $0.09$ & $0.16$ & $\bm{ 0.03 }$ & $0.11$ & {\color{orange} \texttt{Poor}} \\
		forest f & $\bm{ 0.01 }$ & $0.35$ & $0.01$ & $0.01$ & $0.01$ & $0.01$ & $0.01$ & $0.01$ \\
		fps benc & $\bm{ 0.04 }$ & $1.13$ & $4.15$ & $0.36$ & $1.03$ & $0.19$ & $0.25$ & $2.31$ \\
		grid sta & $0.09$ & $80.39$ & {\color{orange} \texttt{Poor}} & $\bm{ 0.03 }$ & $0.18$ & $0.11$ & $0.23$ & $0.15$ \\
		health i & $\bm{ 0.16 }$ & $2.14$ & $59.69$ & $0.25$ & $0.35$ & $0.25$ & $0.30$ & $0.46$ \\
		kin8nm & $1.07$ & $0.85$ & $0.30$ & $0.09$ & $0.08$ & $\bm{ 0.08 }$ & $0.11$ & $0.25$ \\
		kings co & $0.19$ & $0.35$ & $0.38$ & $0.04$ & $0.05$ & $0.03$ & $\bm{ 0.03 }$ & $15.84$ \\
		miami ho & $0.21$ & $0.15$ & $0.16$ & $0.12$ & $\bm{ 0.08 }$ & $0.09$ & $0.08$ & {\color{orange} \texttt{Poor}} \\
		Moneybal & $0.28$ & $6.86$ & $9.48$ & $0.70$ & $\bm{ 0.05 }$ & $4.72$ & $37.20$ & {\color{orange} \texttt{Poor}} \\
		naval pr & $0.09$ & {\color{orange} \texttt{Poor}} & {\color{orange} \texttt{Poor}} & $0.10$ & {\color{orange} \texttt{Poor}} & $0.07$ & $0.42$ & $\bm{ 0.02 }$ \\
		physioch & $0.25$ & $0.21$ & $0.31$ & $\bm{ 0.07 }$ & $0.17$ & $0.19$ & $0.20$ & $0.28$ \\
		pumadyn3 & $0.11$ & {\color{orange} \texttt{Poor}} & $28.13$ & $\bm{ 0.02 }$ & $0.22$ & $0.04$ & $0.04$ & $0.08$ \\
		QSAR fis & $0.64$ & $0.39$ & $28.88$ & $0.86$ & $\bm{ 0.11 }$ & $0.59$ & $0.15$ & $1.50$ \\
		red wine & $0.25$ & $16.32$ & $9.73$ & $2.79$ & $0.09$ & $\bm{ 0.04 }$ & $0.04$ & $0.50$ \\
		sarcos & $0.09$ & $\bm{ 0.04 }$ & $0.07$ & $0.04$ & $0.05$ & $0.27$ & $0.37$ & $0.09$ \\
		socmob & $0.14$ & $\bm{ 0.02 }$ & $0.05$ & $0.21$ & $0.14$ & $0.02$ & $0.08$ & $0.04$ \\
		solar fl & $\bm{ 0.04 }$ & $0.47$ & {\color{orange} \texttt{Poor}} & $0.05$ & $0.55$ & $0.04$ & $0.04$ & $0.37$ \\
		space ga & $0.45$ & $18.07$ & {\color{orange} \texttt{Poor}} & $4.56$ & $0.39$ & $0.09$ & $\bm{ 0.08 }$ & {\color{orange} \texttt{Poor}} \\
		student  & $\bm{ 0.36 }$ & $45.24$ & $8.59$ & $58.97$ & $1.40$ & $2.73$ & $2.07$ & $2.98$ \\
		supercon & $0.07$ & $0.12$ & $0.07$ & $0.11$ & $0.82$ & $\bm{ 0.04 }$ & $0.16$ & $0.16$ \\
		video tr & $0.13$ & $0.41$ & $0.53$ & $0.20$ & $0.08$ & $0.06$ & $0.18$ & $\bm{ 0.04 }$ \\
		wave ene & $0.22$ & {\color{orange} \texttt{Poor}} & $86.37$ & $16.29$ & {\color{orange} \texttt{Poor}} & $\bm{ 0.02 }$ & $0.10$ & $92.01$ \\
		white wi & $0.19$ & $19.54$ & $13.56$ & $3.61$ & $0.14$ & $0.09$ & $\bm{ 0.06 }$ & $1.49$ \\
		\bottomrule
	\end{tabular}
\end{table}

\begin{table}[H]
	\caption{Mean success rate over 5 random seeds by dataset and algorithm.\\}
	\label{tab:success}
	\centering
	\begin{tabular}{lllllllll}
		& Ours & Py410m & Py1b & Ce590m & Ce1.3b & FlanT5 & LMT5 & BART \\
		\midrule
		abalone & $1.00$ & $0.86$ & $0.24$ & $0.15$ & $1.00$ & $1.00$ & $1.00$ & $1.00$ \\
		airfoil  & $1.00$ & $1.00$ & $1.00$ & $1.00$ & $1.00$ & $1.00$ & $1.00$ & $1.00$ \\
		auction  & $1.00$ & $1.00$ & $1.00$ & $0.96$ & $0.87$ & $1.00$ & $1.00$ & $1.00$ \\
		brazilia & $1.00$ & $1.00$ & $1.00$ & $1.00$ & $1.00$ & $1.00$ & $1.00$ & $1.00$ \\
		californ & $1.00$ & $1.00$ & $1.00$ & $1.00$ & $1.00$ & $1.00$ & $1.00$ & $1.00$ \\
		cars & $1.00$ & $1.00$ & $1.00$ & $0.94$ & $0.93$ & $1.00$ & $1.00$ & $1.00$ \\
		concrete & $1.00$ & $1.00$ & $1.00$ & $1.00$ & $0.84$ & $1.00$ & $1.00$ & $1.00$ \\
		cps88wag & $1.00$ & $1.00$ & $1.00$ & $1.00$ & $0.78$ & $1.00$ & $1.00$ & $0.87$ \\
		cpu acti & $1.00$ & $1.00$ & $0.99$ & $0.99$ & $1.00$ & $1.00$ & $1.00$ & $1.00$ \\
		diamonds & $1.00$ & $1.00$ & $1.00$ & $0.99$ & $0.89$ & $1.00$ & $1.00$ & $1.00$ \\
		energy e & $1.00$ & $1.00$ & $1.00$ & $1.00$ & $0.66$ & $1.00$ & $1.00$ & $1.00$ \\
		fifa & $1.00$ & $1.00$ & $1.00$ & $1.00$ & $1.00$ & $1.00$ & $1.00$ & $1.00$ \\
		forest f & $1.00$ & $1.00$ & $1.00$ & $1.00$ & $1.00$ & $1.00$ & $1.00$ & $1.00$ \\
		fps benc & $1.00$ & $0.74$ & $0.21$ & $0.99$ & $0.68$ & $1.00$ & $1.00$ & $0.97$ \\
		grid sta & $1.00$ & $1.00$ & $1.00$ & $1.00$ & $1.00$ & $1.00$ & $1.00$ & $1.00$ \\
		health i & $1.00$ & $0.48$ & $0.13$ & $0.96$ & $1.00$ & $1.00$ & $1.00$ & $1.00$ \\
		kin8nm & $1.00$ & $1.00$ & $1.00$ & $1.00$ & $1.00$ & $1.00$ & $1.00$ & $1.00$ \\
		kings co & $1.00$ & $1.00$ & $1.00$ & $1.00$ & $1.00$ & $1.00$ & $1.00$ & $1.00$ \\
		miami ho & $1.00$ & $1.00$ & $0.95$ & $1.00$ & $0.93$ & $1.00$ & $1.00$ & $1.00$ \\
		Moneybal & $1.00$ & $1.00$ & $1.00$ & $1.00$ & $0.90$ & $1.00$ & $1.00$ & $1.00$ \\
		naval pr & $1.00$ & $1.00$ & $1.00$ & $1.00$ & $1.00$ & $1.00$ & $1.00$ & $0.77$ \\
		physioch & $1.00$ & $1.00$ & $1.00$ & $0.99$ & $1.00$ & $1.00$ & $1.00$ & $1.00$ \\
		pumadyn3 & $1.00$ & $1.00$ & $1.00$ & $1.00$ & $1.00$ & $1.00$ & $1.00$ & $1.00$ \\
		QSAR fis & $1.00$ & $1.00$ & $1.00$ & $0.97$ & $1.00$ & $1.00$ & $1.00$ & $1.00$ \\
		red wine & $1.00$ & $1.00$ & $1.00$ & $0.98$ & $1.00$ & $1.00$ & $1.00$ & $0.82$ \\
		sarcos & $1.00$ & $1.00$ & $0.98$ & $1.00$ & $0.98$ & $1.00$ & $1.00$ & $1.00$ \\
		socmob & $1.00$ & $1.00$ & $1.00$ & $1.00$ & $1.00$ & $1.00$ & $1.00$ & $0.92$ \\
		solar fl & $1.00$ & $1.00$ & $0.19$ & $0.63$ & $0.36$ & $1.00$ & $1.00$ & $1.00$ \\
		space ga & $1.00$ & $1.00$ & $1.00$ & $1.00$ & $0.99$ & $1.00$ & $1.00$ & $1.00$ \\
		student  & $1.00$ & $0.62$ & $0.48$ & $0.16$ & $0.18$ & $1.00$ & $1.00$ & $1.00$ \\
		supercon & $1.00$ & $1.00$ & $1.00$ & $0.74$ & $0.69$ & $1.00$ & $1.00$ & $1.00$ \\
		video tr & $1.00$ & $0.70$ & $0.97$ & $1.00$ & $1.00$ & $1.00$ & $1.00$ & $1.00$ \\
		wave ene & $1.00$ & $1.00$ & $1.00$ & $1.00$ & $1.00$ & $1.00$ & $1.00$ & $1.00$ \\
		white wi & $1.00$ & $1.00$ & $1.00$ & $1.00$ & $1.00$ & $1.00$ & $1.00$ & $0.87$ \\
		\midrule
		mean & $\bm{1.00}$ & $0.95$ & $0.89$ & $0.92$ & $0.90$ & $\bm{1.00}$ & $\bm{1.00}$ & $0.98$ \\
		\bottomrule
	\end{tabular}
\end{table}

\begin{figure}[H]
	\centering
	\includegraphics[angle=90,height=0.95\textheight]{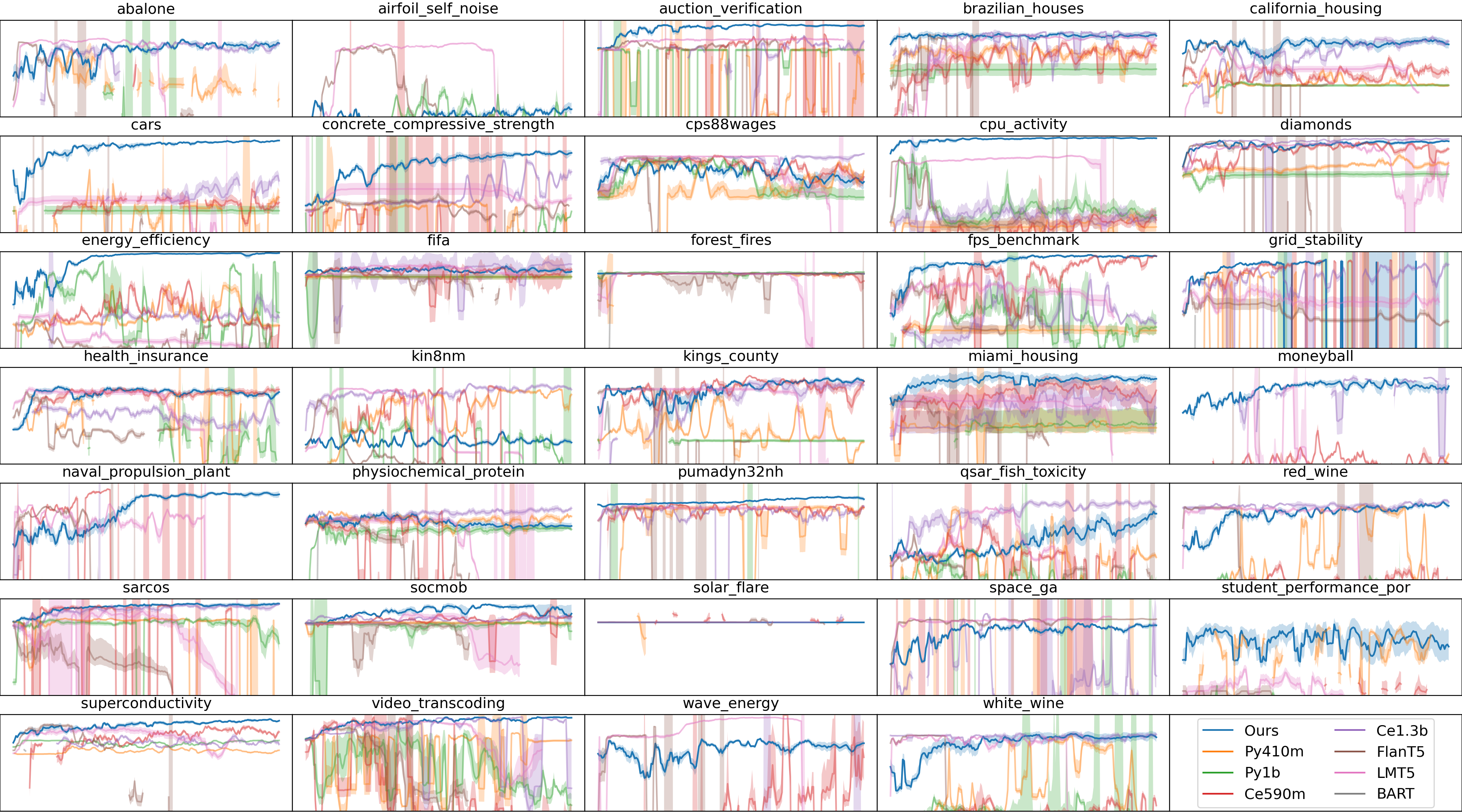}
	\caption{An enlarged view of validation $R^2$ scores over gradient steps per dataset. The shaded area represents the 1 standard deviation across 5 random seeds from the mean. The range of $y$-axis is [-20, 1] (in symmetric log scale) and the range of $x$-axis is [0, 200]. The blue line is from \BT. Higher values are better.}
		\label{fig:evalstepswithspread}
\end{figure}

\begin{table}[H]
	\caption{Run times for each experiment including setup, training, and evaluations.\\}
	\label{tab:runtimes}
	\centering
	\begin{tabular}{llllllll}
		Ours & Py410m & Py1b & Ce590m & Ce1.3b & FlanT5 & LMT5 & BART \\
		\midrule
		21h 35m & 13h 23m & 22h 28m & 36h 49m & 40h 37m & 21h 55m & 21h 00m & 17h 19m\\
		\bottomrule
	\end{tabular}
\end{table}

Note that in \BT, we do not cache the BERT text encoders. We expect better runtimes if we were to cache BERT vectors before running the experiments because our bottleneck is calling the BERT model and padding the input tensors every mini-batch.

\section{Experiment Hyperparameters and Configurations}\label{sec:hyperparams}

\subsection{Datasets}

One dataset (Geographical Origin of Music) was removed because the baselines cannot efficiently handle the large number of columns (68). The size of each of the 34 datasets varies considerably. We want to ensure that the evaluation (validation and test splits) is consistent throughout, and also subject the models to some degree of data imbalance -- a highly common issue for tabular modality. The test and validation set size is 20\% of the number of rows of the smallest dataset, and is randomly sampled; the training set size is the remaining rows.

\subsection{Training Routine}

We do not use the notion of epochs because there are too many samples. Instead, we introduce the notion of \textit{strides}: the training gradient steps before evaluating. There are two quantities of interest: the number of strides (essentially the number of validation passes) and the stride size (number of gradient steps). All models in the multi-task experiment use 200 strides, and a stride size of 200.

\subsection{Basis Transformer}

The text encoder used is a smaller distilled BERT \cite{devlin2019bert}: \texttt{bert\_uncased\_L-8\_H-256\_A-4} by \citet{turc1908well} hosted on Huggingface \cite{wolf2020huggingface}. The reason for using a smaller encoder is because we believe the semantic space across all text in the 34 datasets is relatively small. The embedding dimension of this BERT is 256. The sign-magnitude representation has 29 bits for natural number exponents (i.e. $\{2^{28}, ..., 2^0\}$), 14 bits for negative exponents (i.e. $\{2^{-1}, ..., 2^{-14}\}$) and one sign bit. This was chosen to fit the range of values in the dataset. Five random seeds were used: 0, 1, 2, 3, and 4.

These are the important hyperparameters of \BT:
\begin{itemize}
	\item Embedding dimension: 144
	\item Number of blocks: 4
	\item Number of attention heads: 8
	\item Number of basis queries: 64
	\item $r = 6$
	\item Number of self-attention blocks: 9
	\item Batch size: $64 \times $ number of GPUs
	\item Dropout rate: 0.0
	\item Maximum norm: 1.0
	\item Learning rate: $1 \times 10^{-4}$
	\item Weight decay: $1 \times 10^{-2}$
	\item Exponential decay of learning rate multiplier: 0.985
	\item Adaptive loss reweighing factor $\gamma = 0.2$
\end{itemize}

The optimizer used is the AdamW optimizer with default $\beta$ values. The loss function is binary cross entropy.

\subsection{LLMs with TabLLM Routines}

LLMs face memory (VRAM) issues due to the attention operation across the entire serialized row. As a result, a smaller batch size and gradient accumulation need to be used. Furthermore, since LLMs are pretrained, to be fair to \BT, we reduce the batch size but keep the number of gradient steps the same. LLMs would have already had a good representation learned so it would not need to see that many samples but it still has the same number of finetuning steps as \BT. The finetuning processes using the Huggingface \cite{wolf2020huggingface} interface which offers a convenient method to handle sequence-to-sequence and causal language models.

In general, the batch size is $16 \times $ the number of GPUs, which is often achieved with gradient accumulation. However, the T5 variants consume relatively much more memory than the other LLMs and could only have a batch size of $8 \times $ the number of GPUs. The learning rates and other related hyperparameters are the default or recommended ones listed in Huggingface or their respective papers.

Tabular data often contains a lot of numeric data, and this is definitely true in the OpenML-CTR23 benchmark \cite{fischer2023openml}. If we naively serialize the data, most LLMs would not be able to fit a batch of data into the GPU memory buffer due to the many digits in numeric entry values. To solve this issue while retaining as much relevant information, we round numbers to 5 significant digits. Numbers with less than 5 significant digits are not affected.

Recall that not all LLMs can successfully produce a prediction that fits the template. In cases like this, we compute the $R^2$ scores excluding those failure cases -- in other words, we do not penalize LLMs for such failures. However, in enterprise or important settings, this might be considered a major failure.

\section{Ablation on the Number of \BT Blocks}\label{sec:ablationblocks}

\begin{figure}[H]
	\centering
	\includegraphics[width=0.65\textwidth]{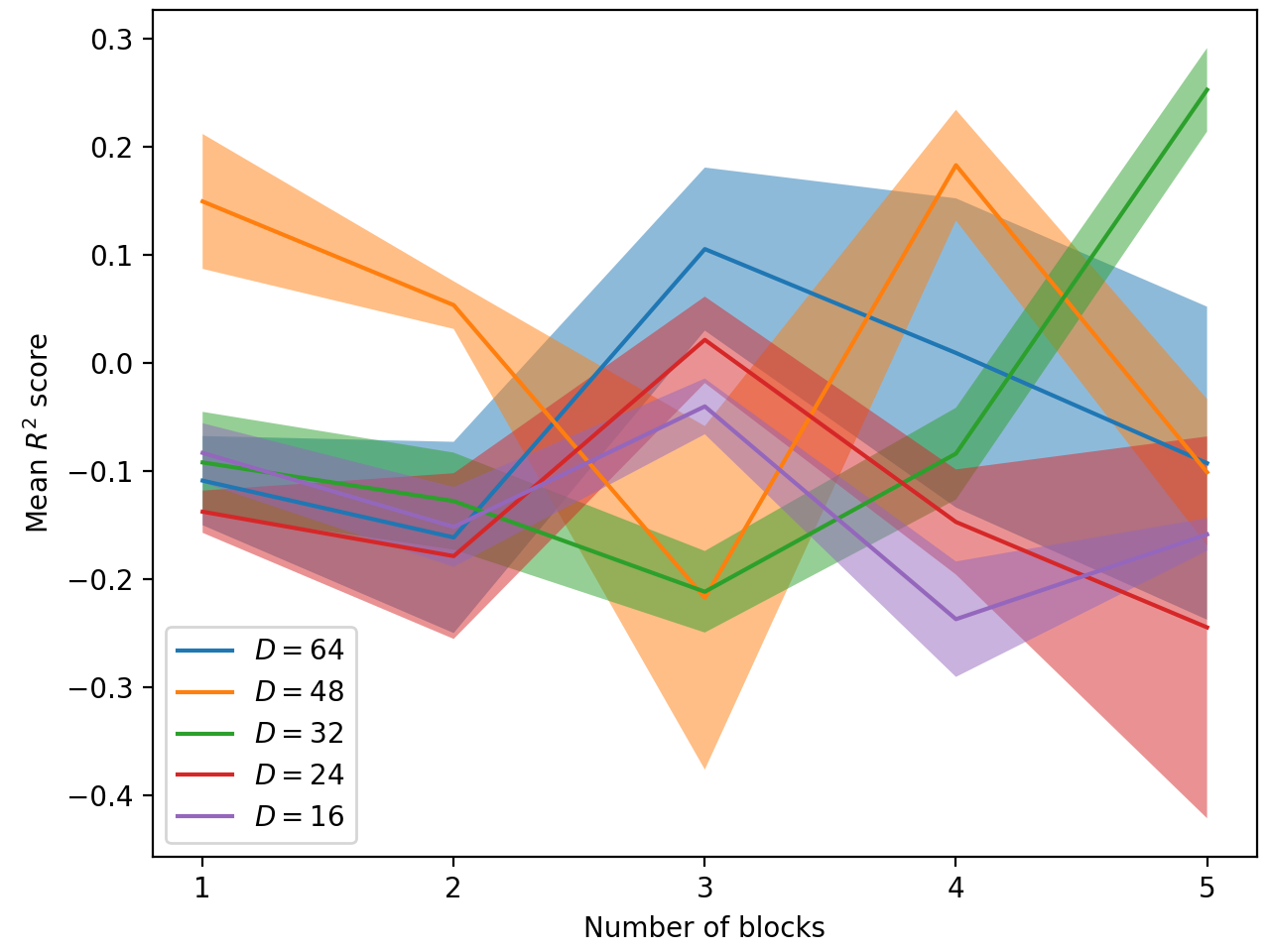}
	\caption{Test performance against the number of \BT blocks over several choices of embedding dimension $D$. The shaded area represents the region of 1 standard error over 5 random seeds. Higher $y$-values are better.}
	\label{fig:blocksablation}
\end{figure}

To show the impact of the number of blocks on a smaller scale experiment, we pick three datasets: abalone, concrete, and white wine. We used a small \BT model:
\begin{itemize}
	\item Number of attention heads: 8
	\item Number of basis queries: 32
	\item $r = 2$
	\item Number of self-attention blocks: 4
	\item Batch size: $128 \times $ number of GPUs
	\item Dropout rate: 0.3
	\item Maximum norm: 1.0
	\item Learning rate: $2 \times 10^{-4}$
	\item Weight decay: $5 \times 10^{-2}$
	\item Exponential decay of learning rate multiplier: 0.99
	\item Adaptive loss reweighing factor $\gamma = 0.45$
\end{itemize}

The number of parameters varies across each experiment due to the changing number of blocks and embedding dimensions. The sign-magnitude representation has 14 bits for natural number exponents (i.e. $\{2^{13}, ..., 2^0\}$), 6 bits for negative exponents (i.e. $\{2^{-1}, ..., 2^{-6}\}$) and one sign bit. We also used a smaller BERT model (bert\_uncased\_L-12\_H-128\_A-2) with an embedding dimension of 128. In Figure \ref{fig:blocksablation}, we show that the best hyperparameters involve more than 1 \BT block. There is no clear relationship in determining the optimal number of blocks.

However, we postulate that the number of blocks, along with the extensive residual connections, can act as a mechanism similar to gradient boosting. Each \BT block can be seen as a weak learner, and the sequential nature of calling the blocks could learn to mitigate the errors of the previous blocks.

\section{Ablation Study on Sign-magnitude and Scalar Outputs}\label{sec:ablationscalar}

\begin{figure}[H]
	\centering
	\includegraphics[width=0.65\textwidth]{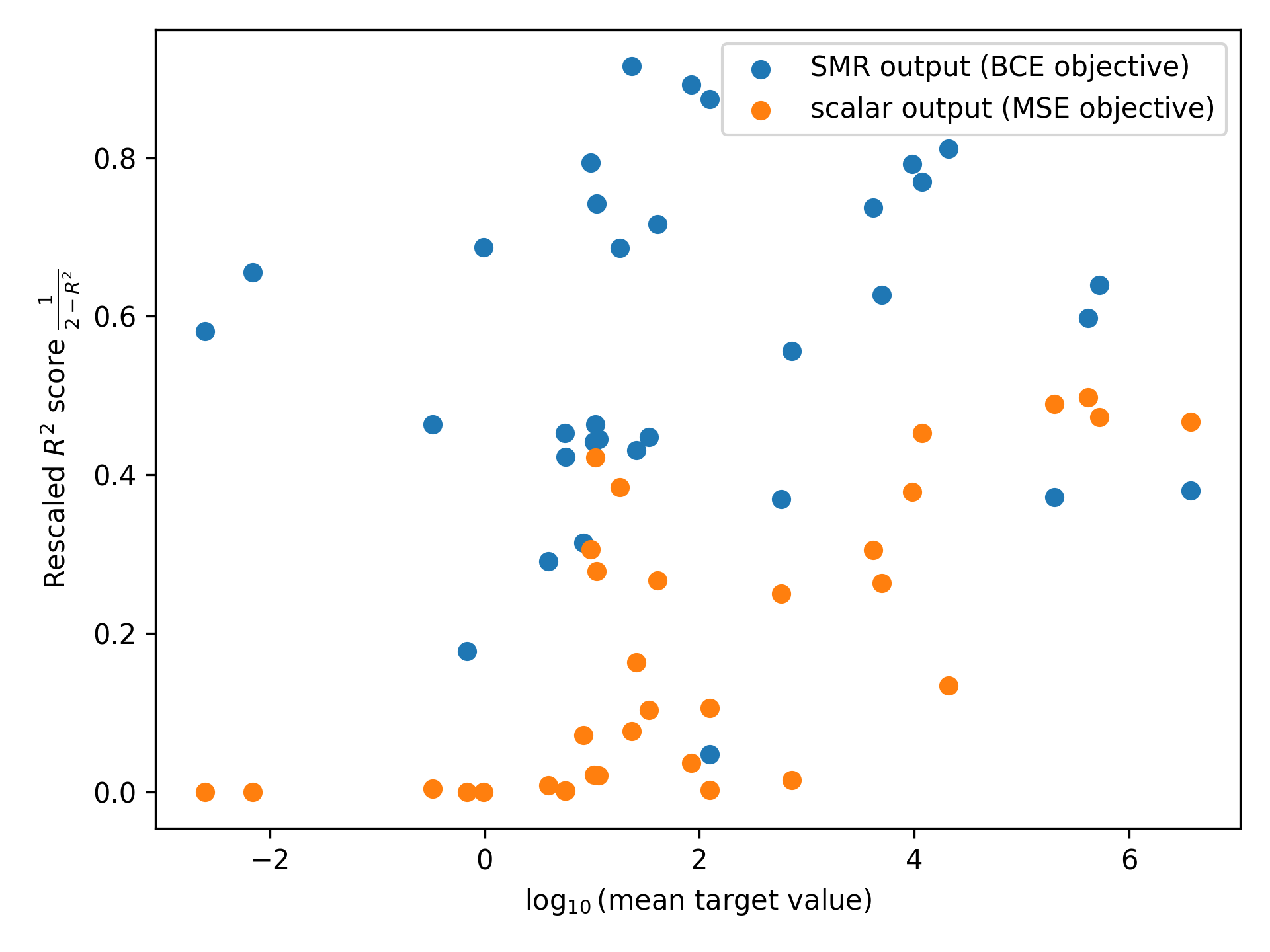}
	\caption{Rescaled test $R^2$ score against log mean target value. Each point represents a dataset, averaged over 5 random seeds. One dataset was excluded because the mean target value is negative valued. Higher $y$-values are better.}
	\label{fig:scalarvsscale}
\end{figure}

We trained an identical model as outlined in Appendix \ref{sec:hyperparams}, except that the output head is replaced such that it produces a single scalar without any activation functions. This model is subjected to the same datasets and training routine, besides using mean squared error instead of binary cross entropy. The results for both are presented in Figure \ref{fig:scalarvsscale}.

The range of $R^2$ is $(-\infty, 1]$, making it difficult to observe any structural patterns. Instead, we rescale it to $(0, 1]$ using the $\frac{1}{2 - R^2}$ transformation. This is the Normalized Nash-Sutcliffe Efficiency (NNSE) metric \cite{mathevet2006bounded}, closely related to $R^2$. The mean target values vary significantly, so we use a log transformation.

Besides the SMR variant being better in raw performance, we can observe a clear relationship between the average target values and the test performance of the scalar variant when trained using mean squared error. Once the mean target value passes $10^{5}$, the scalar variant has competitive performance comparable to the SMR variant but before that, it consistently underperforms.

\section{Ablation Study on Adaptive Loss Reweighing Scheme}\label{sec:ablationadaptiveloss}

\begin{figure}[H]
	\centering
	\includegraphics[width=0.65\textwidth]{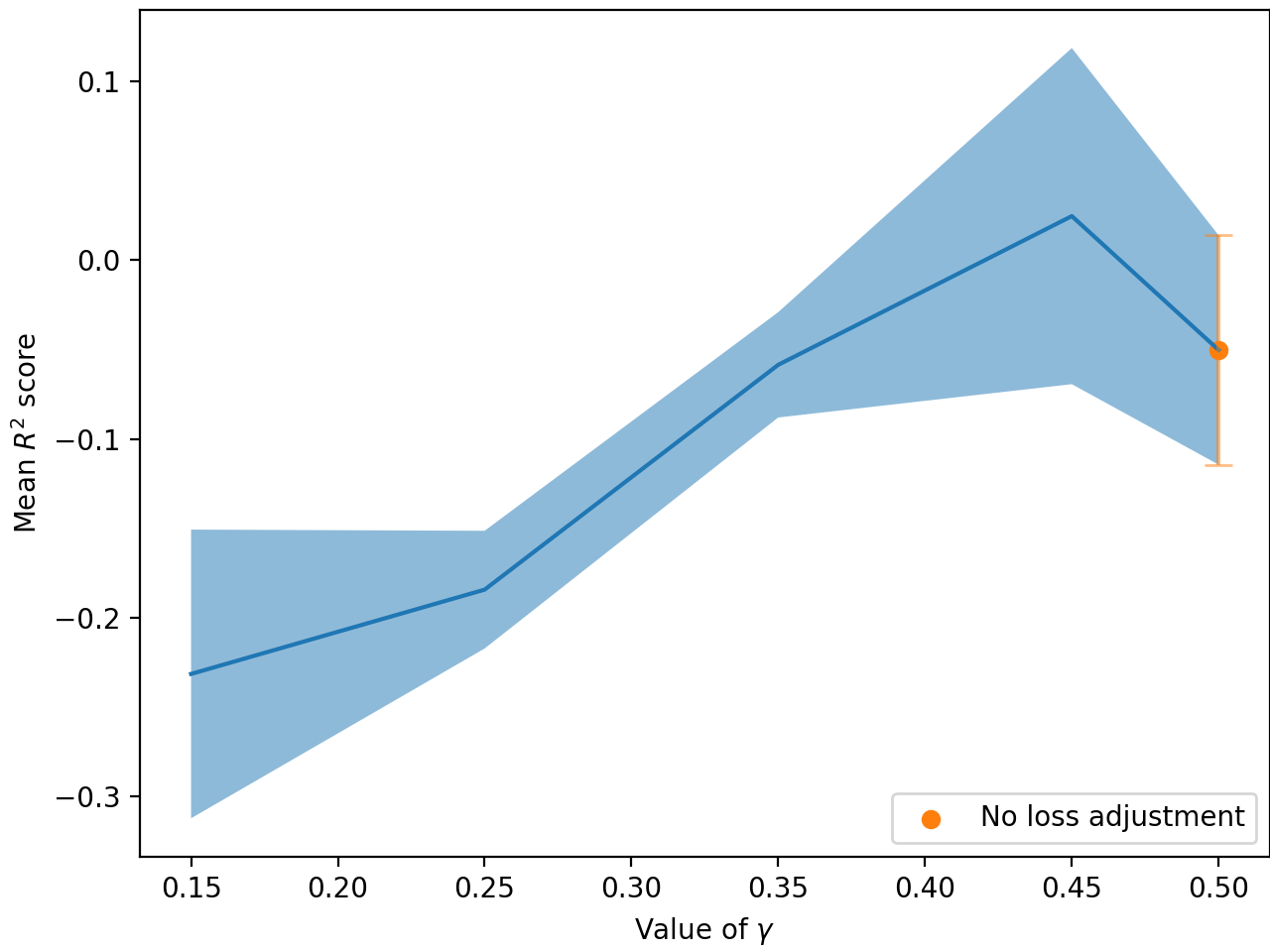}
	\caption{Test performance with and without adaptive loss reweighing scheme over different degrees of reweighing factor $\gamma$. The shaded area represents the region of 1 standard error over 5 random seeds. Higher $y$-values are better.}
	\label{fig:boostingablation}
\end{figure}

To show the efficacy of the reweighing scheme on a smaller scale experiment, we pick three datasets: abalone, concrete, and white wine. We used a small \BT model:
\begin{itemize}
	\item Embedding dimension: 32
	\item Number of blocks: 1
	\item Number of attention heads: 8
	\item Number of basis queries: 96
	\item $r = 2$
	\item Number of self-attention blocks: 5
	\item Batch size: $128 \times $ number of GPUs
	\item Dropout rate: 0.4
	\item Maximum norm: 1.0
	\item Learning rate: $2 \times 10^{-4}$
	\item Weight decay: $5 \times 10^{-2}$
	\item Exponential decay of learning rate multiplier: 0.99
\end{itemize}

The number of parameters is 422,037. The sign-magnitude representation has 14 bits for natural number exponents (i.e. $\{2^{13}, ..., 2^0\}$), 6 bits for negative exponents (i.e. $\{2^{-1}, ..., 2^{-6}\}$) and one sign bit. We also used a smaller BERT model (bert\_uncased\_L-12\_H-128\_A-2) with an embedding dimension of 128. In Figure \ref{fig:boostingablation}, the baseline (no loss reweighing scheme) is the orange point. Recall that $\gamma = 0.5$ means that there are no adjustments; $\gamma \in [0, 0.5)$ means that there is some adjustment, and 0 has the strongest adjustment. We see that with slight adjustments, our model generalized better overall by focusing more on difficult tasks.

While $\gamma = 0.45$ performs the best, in the full experiment, we used a strong adjustment factor of $\gamma = 0.2$ because there are 34 tasks while there are only 3 tasks here.

\section{Hardware Usage}\label{sec:hardware}

The full multi-task experiment used NVIDIA A40 for all model training and evaluation, which was the most powerful GPU we had access to. The time limit was 48 hours. Four GPUs, 32 CPU cores, and 128 GB of RAM were used during the entire process. This was done in an internal cluster with limited details on CPU or RAM.

The ablation studies, however, are done on two NVIDIA Tesla T4 GPUs because the experiments use only 3 datasets and a significantly smaller model. The rest of the hardware configurations are the same.

\end{document}